\documentclass[letterpaper]{article} 
\usepackage{aaai25}  
\usepackage{times}  
\usepackage{helvet}  
\usepackage{courier}  
\usepackage[hyphens]{url}  
\usepackage{graphicx} 
\RequirePackage{subcaption}
\usepackage{multirow}
\usepackage{amsmath}
\usepackage{colortbl}
\usepackage{xcolor}
\usepackage{natbib}  
\usepackage{caption} 
\usepackage{algorithm}
\usepackage{algorithmic}

\usepackage{newfloat}
\usepackage{listings}
\DeclareCaptionStyle{ruled}{labelfont=normalfont,labelsep=colon,strut=off} 
\floatstyle{ruled}
\newfloat{listing}{tb}{lst}{}
\floatname{listing}{Listing}
\title{Texture-AD: An Anomaly Detection Dataset and Benchmark for Real Algorithm Development}
\author{
    Tianwu Lei\textsuperscript{\rm 1}\equalcontrib, 
    Bohan Wang\textsuperscript{\rm 1}\equalcontrib,
    Silin Chen\textsuperscript{\rm 1},
    Shurong Cao\textsuperscript{\rm 1},
    Ningmu Zou\textsuperscript{\rm 1, \rm2}\thanks{Corresponding Author}
}
\affiliations{
    \textsuperscript{\rm 1}School of Integrated Circuits, Nanjing University, Suzhou, China\\
    \textsuperscript{\rm 2}Interdisciplinary Research Center for Future Intelligent Chips (Chip-X), Nanjing University, Suzhou, China \\
    {tianwulei@smail.nju.edu.cn, bohanwang@smail.nju.edu.cn, silin.chen@smail.nju.edu.cn,  shurongcao@smail.nju.edu.cn, nzou@nju.edu.cn}\\
}
\usepackage{bibentry}

\begin{document}
\maketitle

\begin{abstract}
Anomaly detection is a crucial process in industrial manufacturing and has made significant advancements recently.
However, there is a large variance between the data used in the
development and the data collected by the production environment. Therefore, we present the Texture-AD benchmark based
on representative texture-based anomaly detection to evaluate the effectiveness of unsupervised anomaly detection algorithms in real-world applications. This dataset includes images of 15 different cloth, 14 semiconductor wafers and 10 metal plates acquired under different optical schemes. In addition, it includes more than 10 different types of defects produced during real manufacturing processes, such as scratches, wrinkles, color variations and point defects, which are often
more difficult to detect than existing datasets. All anomalous areas are provided with pixel-level annotations to facilitate
comprehensive evaluation using anomaly detection models.
Specifically, to adapt to diverse products in automated pipelines,
we present a new evaluation method and results of baseline
algorithms. The experimental results show that Texture-AD
is a difficult challenge for state-of-the-art algorithms. To
our knowledge, Texture-AD is the first dataset to be devoted to evaluating industrial defect detection algorithms in the real world. The dataset is available at https://XXX.
\end{abstract}

\begin{figure*}[h]
    \centering
    \includegraphics[width=0.9\textwidth]{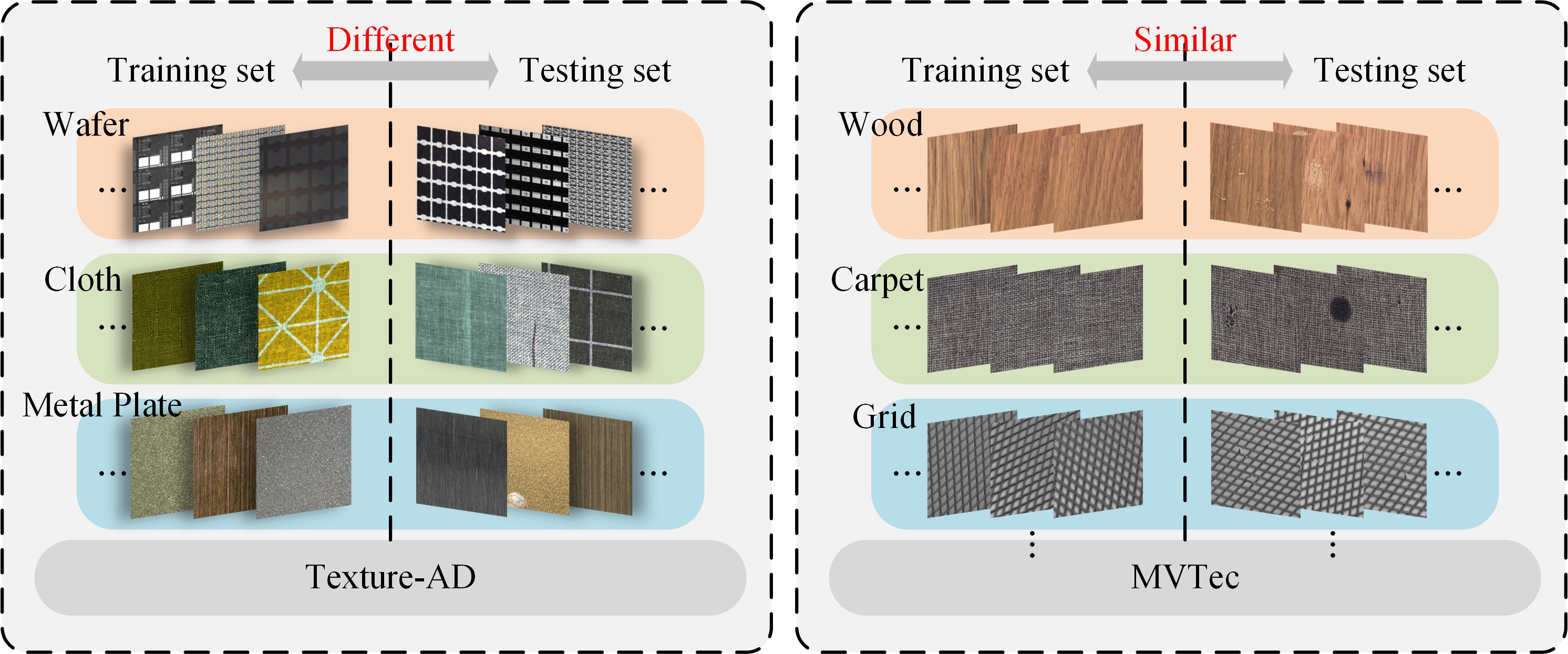}
    \caption{Difference between existing evaluation methods and actual situation}
    \label{fig:1}
\end{figure*}
\begin{table*}[]
\centering
\caption{Evaluation Protocol Difference Between Texture-AD and MVTec}
\begin{tabular}{c|cc|cc}
\hline
\multirow{2}{*}{Category} & \multicolumn{2}{c|}{Train}                    & \multicolumn{2}{c}{Test}                      \\ \cline{2-5} 
                          & \multicolumn{1}{c|}{Images} & Category Labels & \multicolumn{1}{c|}{Images} & Category Labels \\ \hline
MVTec                     & \multicolumn{1}{c|}{O}      & O               & \multicolumn{1}{c|}{O}      & O               \\ \hline
Ours                      & \multicolumn{1}{c|}{O}      & O               & \multicolumn{1}{c|}{O}      & X               \\ \hline
\end{tabular}
\label{table:1}
\end{table*}

\section{Introduction}
Industrial inspection algorithms are typically developed and tested using collected data before deployment, for use in automated quality control equipment on production lines. In recent years, a variety of detection methods have developed for detecting an anomalous image region in image data through contemporary machine learning approaches. These methodologies have demonstrated promising results on established datasets. Present evaluation strategies typically entail integrating flawless production data of a single object category during the training stage and evaluating performance using data containing anomalies.

The acquisition of flawless production data has become more accessible when contrasted with defective data. However, a production line is often required to deal with various specifications of similar products, such as gray cloth, red cloth, mesh cloth, different types of wafers as well as black brushed metal plates, gold frosted metal plates, etc. While these different specifications share certain common features, they also present significant differences. Additionally, minor fluctuations in external conditions, such as lighting environment and camera settings, result in a data distribution after deployment that is unlikely to align with the data collected during the training phase. This situation places increased requirements on the robustness of the algorithms.

Humans have the natural ability to visually discern the similarities and differences in images and to detect defects and irregularities within them. Currently, there are many commonly used datasets for anomaly detection, which vary greatly in the scenes and scale they contain. For example, datasets related to cloth texture\cite{1,2} generally have a good amount of data, but they differ significantly from actual production scenarios. In addition, as chips become an increasingly important field of research worldwide, wafer defect detection has become an essential part of the process. Therefore, the demand for wafer defect detection datasets\cite{3} in industrial inspection is also growing, yet there are very few open-source wafer defect detection datasets available. Moreover, there are more datasets related to metal defects in industrial production\cite{4,5,6,7,8,9}, but they generally include material types and apply to a more limited range of scenarios. There are also datasets related to crack defects\cite{10,11}, such as cracks in bridge surfaces and concrete floors.

So far, modern machine learning systems have encountered considerable challenges in addressing related issues, mainly because the existing datasets are not particularly well-suited to real-world scenarios. Currently, the evaluation of anomaly detection algorithms often relies on datasets such as MVTec\cite{12}, where the features of flawless and defective items show a high degree of consistency, leading to higher performance metrics than actual deployment. Therefore, this paper proposes the Texture-AD dataset\cite{50}, which clearly demonstrates the differences between Texture-AD and the MVTec dataset in Table \ref{table:1}. As shown in Figure \ref{fig:1}, the training data provided by the MVTec dataset and the test data belong to completely the same product, making it impossible to correctly evaluate the algorithms under development. Therefore, in Texture-AD, we provide a variety of specifications of three products as the training set, and at the same time, provide the same type of products with different specifications from the train set as the test set, which can evaluate the performance of the algorithm based on the consideration of algorithm robustness and generalization ability. The training set of this dataset includes $15$ subclasses of cloth images, $14$ subclasses of wafer images and $10$ subclasses of metal plate images. All cloth images come from the same type of cloth, wafer images come from $14$ different subclasses of wafers and metal plate images come from metal plates with $5$ different colors of brushed and matte surfaces, photographed under similar lighting conditions. The test set includes defective cloth images, wafer images and metal plate images photographed from the actual production process, which show slight differences in camera settings, lighting conditions and the design of cloth, wafers and metal plates compared to the training set.

The contributions of our paper can be summarized into three main aspects:

\begin{itemize}
    \item We present a novel and comprehensive dataset for unsupervised anomaly detection in industrial quality inspection. It simulates real-world industrial inspection scenarios and it has a sufficient number of data samples and data scale, including $43120$ high-resolution images collected in various optical environments from $39$ different subclasses under three major categories, which contain a variety of different types of defects.
    \item We conduct a comprehensive evaluation of current state-of-the-art methods for unsupervised anomaly detection, assessing their segmentation and classification performance on the anomalous images during development process.
    \item We provide a well-designed evaluation protocol to compare the performance of unsupervised anomaly detection algorithms in actual development environments.
\end{itemize}

\section{Related Work}
Computer vision equipment for detecting surface defects has largely replaced manual inspections across industries like 3C electronics, automotive, machinery, semiconductors, chemicals and so on. Traditional methods use standard image processing and classifiers with handcrafted features, while effective imaging schemes ensure clear defect visibility under uniform lighting. Recently, deep learning has become prevalent for defect detection. 

DAGM2007 dataset\cite{1} is artificially generated but resembles real-world problems. Six categories referred to as the development dataset, should be used for algorithm development. The remaining four categories (referred to as the competition dataset) can be used to evaluate performance. AITEX dataset\cite{2} is an image dataset focused on the textile industry, designed to support research and application of machine learning and computer vision technology in the field of textile quality inspection. However, the aforementioned two datasets have issues with unclear defect labeling and a rather singular background type and defect type, which cannot fully simulate the complex detection scenarios in actual industrial environments.

The WM-811K dataset\cite{3} is a dataset specifically for semiconductor wafer map defect type identification, with images in the dataset mainly coming from actual production environments of wafer maps, obtained through electrical testing, and used to describe the state of wafer defects. However, the WM-811K represents without texture details and pattern information.

A dataset\cite{4} collected six typical surface defects of hot-rolled steel strips. This surface defect dataset faces two major challenges: large differences in appearance among defects within the same category, and similarities between defects of different categories, with defect images affected by lighting and material changes. The NEU-surface-defect-database\cite{5} has six typical surface defects of hot-rolled steel strips, namely rolling scale, patches, cracks, pitted surfaces, inclusions and scratches. The improved X-SDD dataset\cite{6} includes: seven typical types of hot-rolled steel strip defect images, due to the imbalance of sample quantity in X-SDD, it provides conditions for researchers to solve the problem of sample imbalance. The SD-saliency-900 dataset\cite{7} includes three types of steel strip surface defects (inclusions, patches and scratches), including steel surface defect detection images and corresponding pixel-level binary masks. RSDDS-113 dataset\cite{8}, with samples taken from the actual industrial production line of a section steel factory, collects $20$ track sections with defect information. Each pair of images in this dataset consists of a left camera image and the corresponding depth image; the dataset has a high degree of annotation credibility, but the amount of data samples is fewer. The Rail-5k dataset\cite{9} is used for the task of steel rail surface defect detection. The dataset can be used for two settings, the first is a supervised setting trained with marked images, the fine-grained nature of defect categories and long-tail distribution makes it difficult for visual algorithms to solve. The second is a semi-supervised learning setting promoted by unmarked images, including possible image damage and domain shift with marked images. The dataset can support both supervised and semi-supervised learning settings. In actual production, there may be unknown types of defects, making it difficult for the aforementioned traditional datasets based on known defect patterns to cope. In addition, it is difficult to obtain a large number of defect samples in the aforementioned datasets, leading to the problem of small sample sizes when training deep learning models.

The Concrete Crack Images for Classification dataset\cite{10} is created specifically for the task of concrete crack classification. This dataset typically contains tens of thousands of images of concrete surfaces, showing cracks of different types and severities. The Crack-Detection dataset\cite{11} is designed specifically for crack detection tasks, containing images for training and evaluating crack identification algorithms. These images usually come from various material surfaces, especially concrete and other construction engineering materials, because cracks in these materials may lead to structural problems. The images in the aforementioned datasets have issues with varying quality, including resolution, lighting conditions, angles and background complexity, which may affect the performance of crack detection algorithms in the deployment process.

MVTec\cite{12} contains images of anomalous samples with various defects, manually generated. This is a popular dataset for unsupervised anomaly detection that simulates real-world industrial inspection scenarios. The dataset provides the possibility of evaluating unsupervised anomaly detection methods for various textures and object classes with different types of anomalies. Since it provides pixel-level precise ground truth labels for the abnormal areas in the images, it is possible to evaluate anomaly detection methods for image-level classification and pixel-level segmentation.

In industrial settings, the prevalence of normal samples over defective ones creates a dataset imbalance, affecting model training and generalization. Acquiring a significant number of defective samples is costly and time-consuming, especially for rare defects. Current datasets may not cover all defect types, limiting the model's ability to identify unusual defects. The complexity of industrial products' appearance and potential labeling inconsistencies add to the challenge of defect detection. Moreover, the need for real-time responses in industry is often not met by existing datasets, leading to models that may not perform well in new environments.

\begin{figure*}[htbp]
	\centering
	\begin{subfigure}{0.33\linewidth}
		\centering
		\includegraphics[width=1\linewidth]{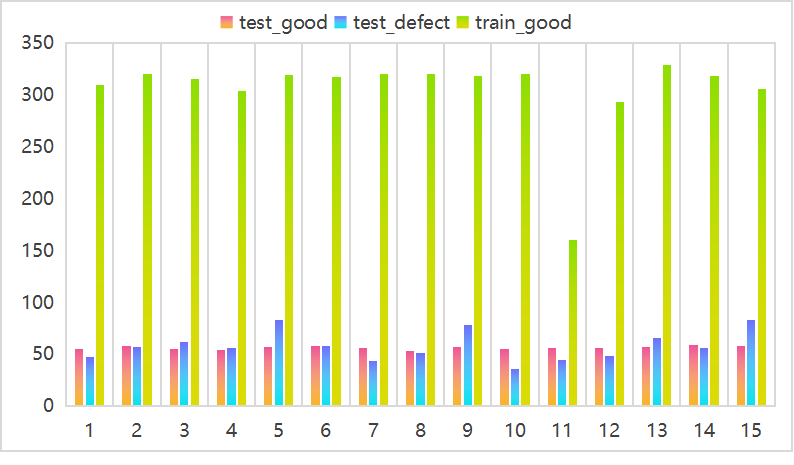}
		\caption{Cloth dataset}
	\end{subfigure}
	\centering
	\begin{subfigure}{0.33\linewidth}
		\centering
		\includegraphics[width=1\linewidth]{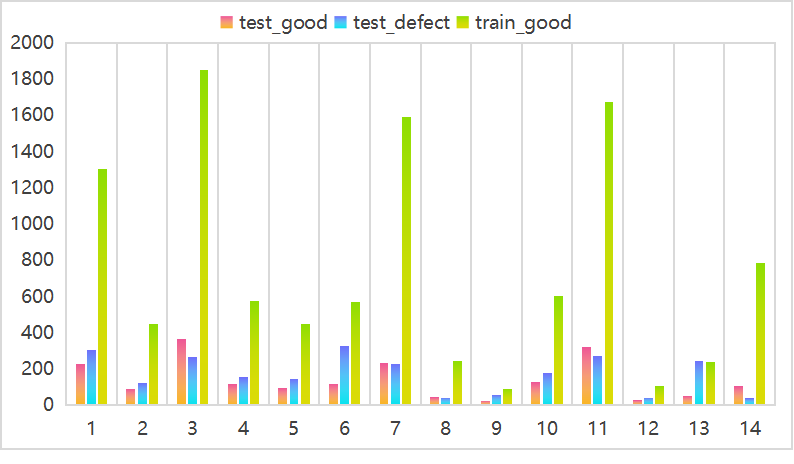}
		\caption{Wafer dataset}
	\end{subfigure}
	\centering
	\begin{subfigure}{0.33\linewidth}
		\centering
		\includegraphics[width=1\linewidth]{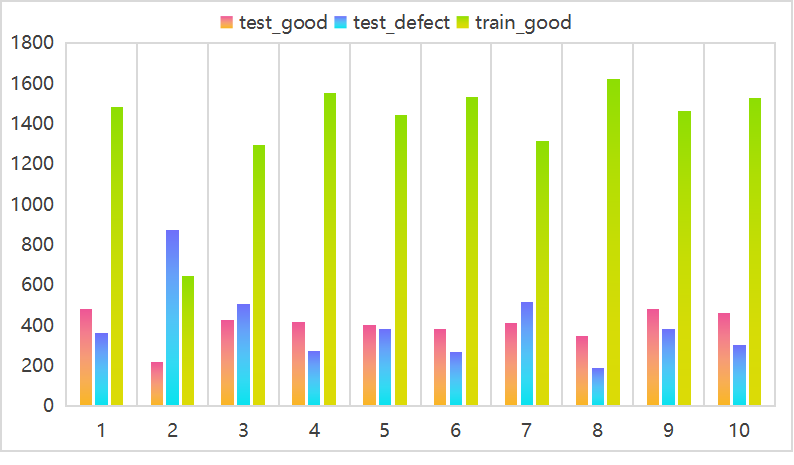}
		\caption{Metal plate dataset}
	\end{subfigure}
 \caption{Data Statistics (a)The cloth dataset consists of a total of $6283$ images, with $4569$ images in the training set and $1714$ images in the test set. (b)The wafer dataset consists of a total of $14861$ images, with $10525$ images in the training set and $4336$ images in the test set. (c)The metal plate dataset consists of a total of $21976$ images, with $13879$ images in the training set and $8097$ images in the test set. }
	\label{fig:2}
\end{figure*}
\begin{figure}[h]
    \centering
    \includegraphics[width=0.45\textwidth]{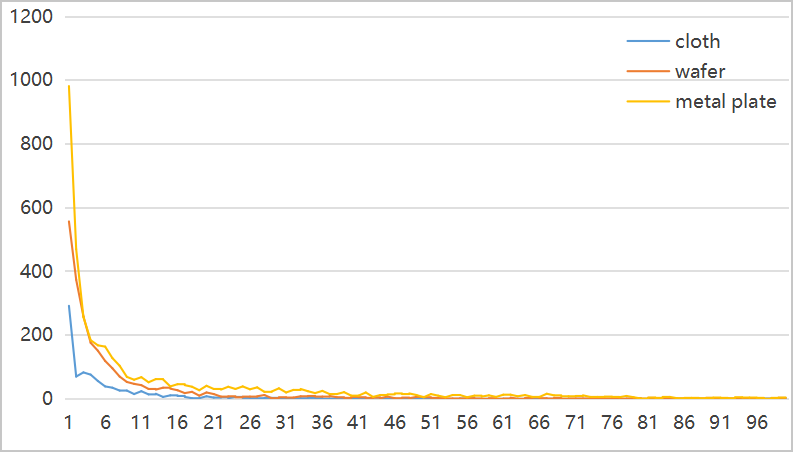}
    \caption{Statistics of the percentage of the image area occupied by the anomaly region}
    \label{fig:3}
\end{figure}
\begin{figure*}[h]
    \centering
    \includegraphics[width=0.9\textwidth]{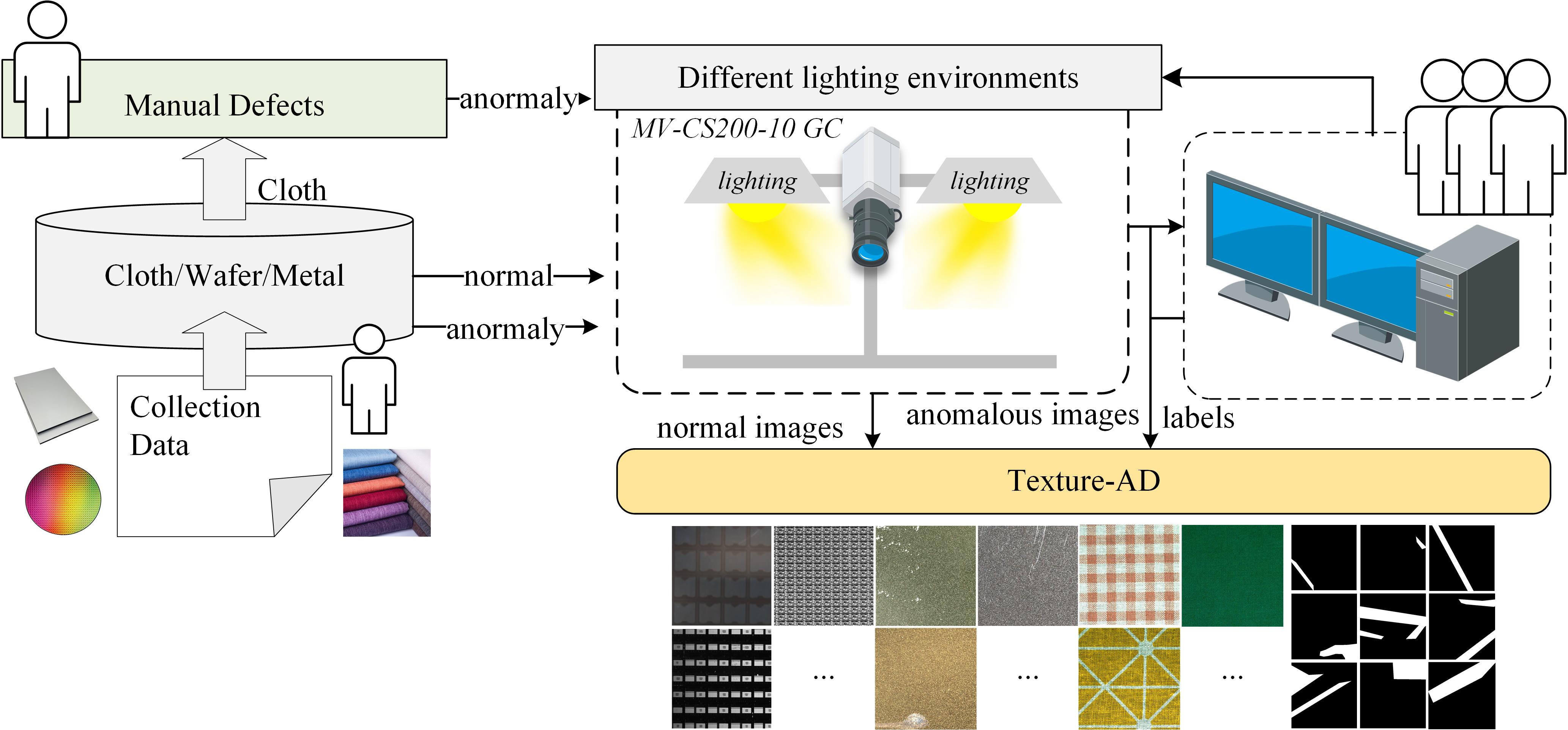}
    \caption{Image acquisition and defect annotation processes. The Texture-AD images were captured using a high-resolution industrial camera (MV-CS200-10 GC). The optical scheme was altered by adjusting the position of the light source and the brightness of two light sources. The cloth images include both artificial and natural defects, while the wafer and metal plate images consist solely of natural defects. The defect annotation work for the images was performed using Labelme.}
    \label{fig:4}
\end{figure*}

\section{Dataset}
The anomaly detection dataset we propose includes $15$ subclasses of cloth, covering a variety of colors, materials and texture defects, $14$ different subclasses of wafers and $10$ subclasses of metal plates, including $5$ colors each with brushed and matte finishes, totaling $10$ subclasses of textures. The defects in our dataset are imperfections that occur in actual production environments, making it extremely valuable for the study of industrial quality inspection algorithms. Cloth defects include pencil marks, cuts, marker stains, water stains, black and white dots, threads, inconsistent sewing distances and color differences caused by dyeing. Wafer and metal plate defects include scratches, stains and inherent manufacturing defects, all of which naturally occur in the production process. As shown in Figure \ref{fig:2}, Our dataset contains a total of $43120$ images, with $28973$ images used for training and validation, and $14147$ images for testing. The training set includes only defect-free images. The test set contains two types of images: images with various types of defects and defect-free images. Figure \ref{fig:3} shows the percentage of the image area occupied by the anomalous regions.

Specific to the division of the dataset, we provide good production images from multiple subclasses for each category as the training set, allowing the model to learn the characteristics and differences of each subclass. At the same time, we also provide defect images and good production images from the same category for the test set to evaluate the model's recognition ability when facing actual defects. The number of samples for each category and the specific allocation of subclasses are detailed in the appendix for reference.

\subsection{Data Generation}
All images were captured using a high-resolution industrial camera 
(MV-CS200-10 GC) at a resolution of $5472 \times 3648$ pixels, 
in conjunction with two light sources. The optical scheme was altered by adjusting the position and brightness of the light sources. Our image acquisition and defect annotation process is depicted in Figure \ref{fig:4}. The defects in our dataset were manually annotated using the Labelme annotation tool. To better align with the defects produced in the industrial manufacturing process, we created some artificial defects on the cloth, while the wafers and metal plates exhibited naturally occurring defects. Subsequently, these images were cropped to the appropriate output size. All images have a resolution of $1024 \times 1024$ pixels. The training set images were obtained under relatively stable lighting conditions. However, for the test set, we intentionally varied the optical scheme to simulate the imaging discrepancies between the algorithm training phase and actual deployment. We provided pixel-level ground truth annotations for each defective image area. The specific quantities for each category are listed in Figure \ref{fig:4}.  
\section{Anomaly Detection Methods}
The current research trend in anomaly detection is primarily focused on unsupervised anomaly detection. This trend has emerged due to the fact that obtaining anomalous samples requires a significant investment of human and financial resources. In this research context, training data contains only normal samples, while test data includes both normal and anomalous samples. Industrial image anomaly detection is a specific branch within the field of anomaly detection, and we mainly evaluate and compare it using the following three research directions.
\subsection{Synthesis-based Anomaly Detection}
Some supervised learning methods use a limited number of anomaly samples to synthesize more anomaly samples to enhance training effectiveness. For example, A basic architecture that integrates CycleGAN\cite{14} with ResNet/U-Net as the generator is used to transfer defects from one image to another\cite{13}. SDGAN\cite{15} achieved better results than CycleGAN by improving the style transfer network. DRAEM\cite{16} first restores the normal image with pseudo-anomaly interference to obtain feature representation and then uses a discriminator network to distinguish anomalies, demonstrating excellent performance. Although this field has made certain research progress, it still has a huge development space compared to other fields with clear research directions.
\subsection{Reconstruction-based Anomaly Detection}
These methods are based on the assumption that a reconstruction model trained only on normal samples can successfully reconstruct images in normal areas\cite{17,18,19,20,21} but fail in abnormal areas. Early attempts included autoencoders(AE)\cite{17,22}, variational autoencoders(VAE)\cite{19,23} and generative adversarial networks(GAN)\cite{20,24,25,26}. However, these methods may cause the model to learn certain tricks, leading to the effective recovery of anomalies as well. To address this issue, researchers have adopted various strategies, such as introducing guidance information (structure\cite{27} or semantics\cite{28,29}), memory mechanisms\cite{30,31,32}, iterative mechanisms\cite{33}, image masking strategies\cite{34} and pseudo-anomaly\cite{22,35}.PyramidFlow\cite{47} based on the transformer and further design set a new record on MVTec.
\subsection{Feature-Embedding Based Methods}
Feature embedding methods are committed to distinguishing normal and abnormal samples at the feature representation level. Uniformed Students\cite{36} pioneered the use of discriminative latent embeddings for anomaly detection. This model is simple and effective, significantly outperforming other benchmark methods. STPM\cite{37} and MKD\cite{38} utilize multi-scale features on different network layers for feature distillation, although there are differences in their methods. In addition, SimpleNet\cite{39} has achieved satisfactory results by introducing noise into the feature embedding to simulate negative samples.
\begin{table*}[]
\centering
\caption{Comparison of state-of-the-art works on the cloth of Texture-AD.Image-AUROC (top row) and Pixel-AUROC(bottom row) are
 displayed in each entry.}
\label{table:2}
\begin{tabular}{c|cccccc}
\hline
Category                     & subclass1      & subclass2      & subclass3      & subclass4      & subclass5      & Average        \\ \hline
\multirow{2}{*}{SimpleNet}   & 65.08          & 59.26          & 58.83          & \textbf{70.40} & 68.47          & \textbf{64.41} \\
                             & 58.30          & 51.52          & \textbf{63.48} & \textbf{70.68} & 54.47          & 59.69          \\ \hline
\multirow{2}{*}{PyramidFlow} & 57.88          & 63.18          & 60.74          & 59.39          & 49.72          & 58.18          \\
                             & \textbf{68.00} & 57.06          & 60.74          & 57.26          & 34.84          & 55.58          \\ \hline
\multirow{2}{*}{Mean-Shift}  & \textbf{66.22} & 33.66          & \textbf{66.21} & 65.69          & 39.54          & 54.26          \\
                             & -              & -              & -              & -              & -              & -              \\ \hline
\multirow{2}{*}{DRAEM}       & 57.58          & 50.21          & 55.44          & 58.01          & 55.95          & 55.44          \\
                             & 60.99          & \textbf{65.36} & 56.91          & 53.45          & \textbf{77.03} & \textbf{62.75} \\ \hline
\multirow{2}{*}{MSFlow}      & 50.00          & 54.01          & 50.00          & 50.00          & 50.14          & 50.83          \\
                             & 56.11          & 63.14          & 51.66          & 47.44          & 42.23          & 52.12          \\ \hline
\multirow{2}{*}{EfficientAD} & 65.65          & \textbf{76.98} & 55.69          & 42.38          & \textbf{72.20} & 62.58          \\
                             & 62.76          & 58.92          & 47.08          & 38.75          & 61.77          & 53.86          \\ \hline
\end{tabular}
\end{table*}
\begin{table}[]
\centering
\caption{Comparison of state-of-the-art works on the wafer of Texture-AD.Image-AUROC (top row) and Pixel-AUROC(bottom row) are
 displayed in each entry.}
\label{table:3}
\setlength{\tabcolsep}{0.01mm}{}
\begin{tabular}{c|ccccc}
\hline
Category                     & subclass1      & subclass2      & subclass3      & subclass4      & Average        \\ \hline
\multirow{2}{*}{SimpleNet}   & 52.11          & \textbf{59.66} & 53.66          & 50.68          & 54.03          \\
                             & \textbf{57.18} & \textbf{66.16} & \textbf{57.58} & \textbf{53.40} & \textbf{58.58} \\ \hline
\multirow{2}{*}{PyramidFlow} & 55.54          & 43.35          & 52.76          & 46.36          & 49.50          \\
                             & 51.23          & 39.47          & 51.52          & 44.63          & 46.71          \\ \hline
\multirow{2}{*}{Mean-Shift}  & 52.83          & 53.29          & 55.44          & 48.28          & 52.47          \\
                             & -              & -              & -              & -              & -              \\ \hline
\multirow{2}{*}{DRAEM}       & \textbf{55.69} & 57.09          & \textbf{59.22} & \textbf{52.46} & \textbf{56.12} \\
                             & 44.91          & 34.10          & 35.01          & 43.59          & 39.40          \\ \hline
\multirow{2}{*}{MSFlow}      & 51.19          & 49.78          & 53.64          & 50.00          & 51.15          \\
                             & 44.91          & 34.10          & 35.01          & 43.59          & 39.40          \\ \hline
\multirow{2}{*}{EfficientAD} & 50.28          & 42.25          & 50.23          & 45.51          & 47.07          \\
                             & 55.76          & 33.98          & 51.53          & 40.02          & 45.32          \\ \hline
\end{tabular}
\end{table}
\begin{table}[]
\centering
\caption{Comparison of state-of-the-art works on the metal plate of Texture-AD. Image-AUROC (top row) and Pixel-AUROC(bottom row) are
 displayed in each entry.}
\label{table:4}
\setlength{\tabcolsep}{1.3mm}{}
\begin{tabular}{c|cccc}
\hline
Category                     & subclass1      & subclass2      & subclass3      & Average        \\ \hline
\multirow{2}{*}{SimpleNet}   & 59.07          & \textbf{59.87} & 57.83          & 58.92          \\
                             & 62.27          & \textbf{58.33} & 58.97          & 59.86          \\ \hline
\multirow{2}{*}{PyramidFlow} & 52.87          & 48.74          & 58.92          & 53.51          \\
                             & 53.42          & 48.86          & 57.67          & 53.31          \\ \hline
\multirow{2}{*}{Mean-Shift}  & 44.34          & 47.39          & 45.04          & 53.29          \\
                             & -              & -              & -              & -              \\ \hline
\multirow{2}{*}{DRAEM}       & 52.07          & 56.32          & 51.48          & 45.59          \\
                             & 58.41          & 51.53          & 57.31          & 55.75          \\ \hline
\multirow{2}{*}{MSFlow}      & 62.90          & 53.54          & 59.78          & 58.74          \\
                             & \textbf{65.37} & 57.34          & \textbf{60.37} & \textbf{61.02} \\ \hline
\multirow{2}{*}{EfficientAD} & \textbf{65.27} & 55.46          & \textbf{68.73} & \textbf{63.30} \\
                             & 59.69          & 51.04          & 54.91          & 55.21          \\ \hline
\end{tabular}
\end{table}

\section{Benchmark}
\subsection{Baseline Methods}

\subsubsection{SimpleNet}
SimpleNet\cite{39} proposed a simple and easy-to-apply network for detecting and localizing anomalies in images. We evaluated using the publicly available SimpleNet implementation on Pytorch. The backbone network used Wide Resnet50 as the backbone network, setting the feature dimension of the feature extractor to $1536$ to accommodate $329\times329$ sized input images. The anomaly feature generator added isotropic Gaussian noise $N(0, \sigma^2)$, where $\sigma$ defaults to $0.015$. The subsequent discriminator includes a linear layer, batch normalization layer, leaky ReLU with a slope of $0.2$ and a linear layer. The Adam optimizer was used, with learning rates of $0.0001$ and $0.0002$ set for the feature adapter and discriminator, respectively and a weight decay of $0.00001$. Each dataset was trained for $160$ epochs with a batch size of $8$.

\subsubsection{PyramidFlow} 
PyramidFlow\cite{47} proposed a new anomaly localization method, which is based on the defect contrastive localization paradigm using a pyramid of normalization flows for multi-scale fusion and volume normalization to achieve high-resolution defect localization. We used a fixed pyramid layer number $L=8$, image resolution of $256 \times 256$ and channel number $C=24$, and varied the stacked layer number $D$ to explore the trends in memory usage and model parameterization. During training, sample mean normalization was used, and the running mean was updated with a momentum of $0.1$. At test time, volume normalization was based on the running mean.

\subsubsection{Mean-Shift}
Mean-Shift\cite{46} introduced a novel self-supervised representation learning method to improve anomaly detection. It pointed out that traditional contrastive learning methods are not suitable for pre-trained features, hence they proposed the Mean-Shifted Contrastive Loss. In the experiments targeting ResNet152, we fine-tuned the last two blocks of a ResNet152 model pre-trained on the ImageNet dataset and added an $\ell_2$  normalization layer, a process that lasted for $10$ training epochs. For the experiments with ResNet18, we fine-tuned the entire backbone of a ResNet18 model pre-trained on ImageNet and similarly added an $\ell_2$  normalization layer, a process that included 20 training epochs. In both cases, we minimized the Mean-Shifted Contrastive loss function with a temperature parameter $\tau$ set to $0.25$. We used the Stochastic Gradient Descent (SGD) optimizer with a weight decay of $5 \times 10^{-5}$, and without momentum. We set the size of each mini-batch to $64$.
\begin{figure}[h]
    \centering
    \includegraphics[width=0.45\textwidth]{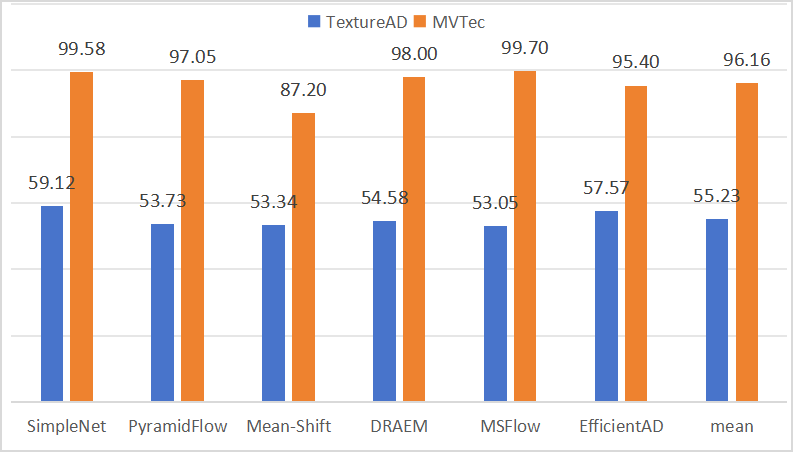}
    \caption{The comparison of the average Image-AUROC obtained by various algorithms on Texture-AD and MVTec}
    \label{fig:5}
\end{figure}
\begin{figure*}[h]
    \centering
    \includegraphics[width=0.7\textwidth]{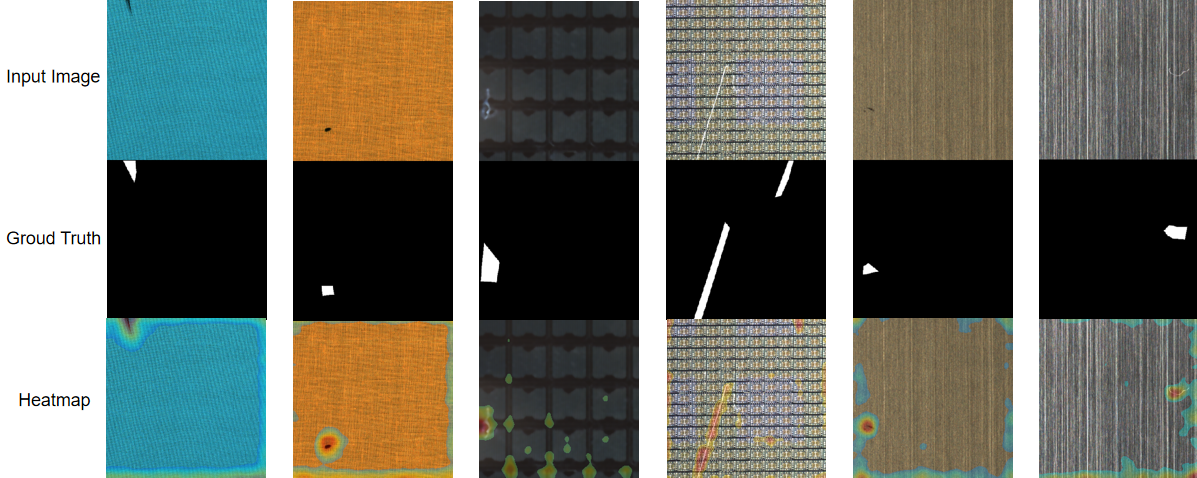}
    \caption{Visualization of SimpleNet results. It presents the anomaly segmentation results for three categories of materials in Texture-AD: cloth, wafer and metal plate. The top row demonstrates the origin image, the medium row shows pixel defect region annotation, and the bottom row is the heatmap of SimpleNet.}
    \label{fig:6}
\end{figure*}
\subsubsection{DRAEM}
In addition to reconstruction methods, DRAEM\cite{16} primarily regards surface anomaly detection as a discriminative problem and proposes a Discriminatively Trained Reconstruction Anomaly Embedding Model (DRAEM). This method learns the joint representation of anomalous images and their anomaly-free reconstructions while learning the decision boundary between normal and anomalous examples. The method can directly localize anomalies without the need for additional complex post-processing of the network output and can be trained using simple and universal anomaly simulation. In our experiments, the network was trained for $700$ epochs. The learning rate was set to $10^{-4}$, and it was multiplied by $0.1$ after $400$ and $600$ epochs. Image rotation from $-45$ to $45$ degrees was used as a data augmentation method.

\subsubsection{MSFlow}
MSFlow\cite{49} proposed a multi-scale flow-based framework for unsupervised anomaly detection, which utilizes normalization flows to handle features at different scales to adapt to anomalies of various sizes. During the experimental process, we used Wide ResNet50 and ResNet18 as feature extractors. The training was conducted with a batch size of $16$. The optimizer used was Adam with an initial learning rate of$10^{-4}$, and the learning rate was reduced at $70\%$ and $90\%$ of the training progress.

\subsubsection{EfficientAD}
EfficientAD\cite{48} proposed a lightweight feature extractor that processes images with millisecond-level latency on modern GPUs, using a student-teacher approach to detect anomalous features and effectively detect logical anomalies. In the experiments, we set the hard feature loss mining factor ($\text{phard}$) to $0.999$, meaning that on average, $10\%$ of the values in each dimension are used for backpropagation. The Adam optimizer was used with an initial learning rate of $10^{-4}$ and a weight decay of $10^{-5}$. During training, if the number of iterations exceeded $66500$, the learning rate was reduced to $10^{-5}$.

\subsection{Evaluation Method}
\subsubsection{Train and Test data }As shown in Table \ref{table:1}, the information available during the training process is the same as for MVTec, but the sub-category labels cannot be used during the testing process.
\subsubsection{Data Augmentation }Since the evaluated methods based on deep learning are typically trained on large datasets, data augmentation is performed for these methods for both textures and objects. We resize the image to fit the shape of the model input.  Additional mirroring is applied. We augment each category to create $10000$ training images.
\subsubsection{Evaluation Metric }Following prior works\cite{12,26,36}, the Area Under the Receiver Operating Curve (AUROC)is used as the evaluation metric for anomaly detection. Image-level anomaly detection performance is measured via the standard Area Under the Receiver Operator Curve, which we denote as I-AUROC. For anomaly localization, we use an evaluation of pixel-wise AUROC (denoted as P-AUROC).

\subsection{Result}
As shown in Table \ref{table:2}, Table \ref{table:3} and Table \ref{table:4}, we present the evaluation results of anomaly image classification and anomaly region segmentation for all methods and dataset categories, respectively. No method performs consistently well across all texture categories. In the cloth category, SimpleNet outperforms the other methods. But in the wafer category, DRAEM performs better than SimpleNet. In the metal plate category, EfficientAD leads the second place by 4.38\% in I-AUROC. 
As shown in Figure \ref{fig:5}, when applying our dataset Texture-AD for evaluation alongside the MVTec dataset, it was found that the evaluation results of our dataset are generally lower, which can expose the problem domains where the algorithm fails, facilitating targeted optimization of the algorithm's weak points in subsequent improvements. Here are the evaluation results of each method. Some examples of performance were provided.(Figure \ref{fig:6}).All experimental results are the mean of
3 replicates.
\subsection{Conclusion}
We introduce the Texture-AD Anomaly Detection Benchmark, a novel dataset for unsupervised anomaly detection that mimics real-world industrial detection scenarios. The dataset provides a way to evaluate unsupervised anomaly detection methods in realistic algorithm development scenarios. Since pixel-accurate ground truth labels of anomaly regions in images are provided, both image-level classification and pixel-level segmentation anomaly detection methods can be evaluated. Several state-of-the-art methods are evaluated on this dataset. The evaluation provided a benchmark for showing how different algorithms perform in real-world application scenarios and indicating that there is still much room for improvement. We hope that the proposed dataset will stimulate the development of new unsupervised anomaly detection methods.

\bibliography{aaai25}

\begin{thebibliography}{44}
\providecommand{\natexlab}[1]{#1}

\bibitem[{Akcay, Atapour-Abarghouei, and Breckon(2018)}]{24}
Akcay, S.; Atapour-Abarghouei, A.; and Breckon, T.~P. 2018.
\newblock GANomaly: Semi-Supervised Anomaly Detection via Adversarial Training.
\newblock arXiv:1805.06725.

\bibitem[{Bao et~al.(2021)Bao, Song, Liu, Wang, Yan, Yu, and Li}]{4}
Bao, Y.; Song, K.; Liu, J.; Wang, Y.; Yan, Y.; Yu, H.; and Li, X. 2021.
\newblock Triplet-Graph Reasoning Network for Few-Shot Metal Generic Surface Defect Segmentation.
\newblock \emph{IEEE Transactions on Instrumentation and Measurement}, 70: 1--11.

\bibitem[{Batzner, Heckler, and König(2024)}]{48}
Batzner, K.; Heckler, L.; and König, R. 2024.
\newblock EfficientAD: Accurate Visual Anomaly Detection at Millisecond-Level Latencies.
\newblock arXiv:2303.14535.

\bibitem[{Bergmann et~al.(2019)Bergmann, Fauser, Sattlegger, and Steger}]{12}
Bergmann, P.; Fauser, M.; Sattlegger, D.; and Steger, C. 2019.
\newblock MVTec AD — A Comprehensive Real-World Dataset for Unsupervised Anomaly Detection.
\newblock In \emph{2019 IEEE/CVF Conference on Computer Vision and Pattern Recognition (CVPR)}, 9584--9592.

\bibitem[{Bergmann et~al.(2020)Bergmann, Fauser, Sattlegger, and Steger}]{36}
Bergmann, P.; Fauser, M.; Sattlegger, D.; and Steger, C. 2020.
\newblock Uninformed Students: Student-Teacher Anomaly Detection With Discriminative Latent Embeddings.
\newblock In \emph{2020 IEEE/CVF Conference on Computer Vision and Pattern Recognition (CVPR)}. IEEE.

\bibitem[{Bergmann et~al.(2018)Bergmann, L{\"{o}}we, Fauser, Sattlegger, and Steger}]{17}
Bergmann, P.; L{\"{o}}we, S.; Fauser, M.; Sattlegger, D.; and Steger, C. 2018.
\newblock Improving Unsupervised Defect Segmentation by Applying Structural Similarity to Autoencoders.
\newblock \emph{CoRR}, abs/1807.02011.

\bibitem[{Chen et~al.(2022)Chen, You, Zhang, Xi, and Le}]{18}
Chen, L.; You, Z.; Zhang, N.; Xi, J.; and Le, X. 2022.
\newblock UTRAD: Anomaly detection and localization with U-Transformer.
\newblock \emph{Neural Networks}, 147: 53--62.

\bibitem[{Chu, Zhmoginov, and Sandler(2017)}]{14}
Chu, C.; Zhmoginov, A.; and Sandler, M. 2017.
\newblock CycleGAN, a Master of Steganography.
\newblock \emph{CoRR}, abs/1712.02950.

\bibitem[{Collin and De~Vleeschouwer(2021)}]{22}
Collin, A.-S.; and De~Vleeschouwer, C. 2021.
\newblock Improved anomaly detection by training an autoencoder with skip connections on images corrupted with Stain-shaped noise.
\newblock In \emph{2020 25th International Conference on Pattern Recognition (ICPR)}, 7915--7922.

\bibitem[{Dehaene et~al.(2020)Dehaene, Frigo, Combrexelle, and Eline}]{33}
Dehaene, D.; Frigo, O.; Combrexelle, S.; and Eline, P. 2020.
\newblock Iterative energy-based projection on a normal data manifold for anomaly localization.
\newblock arXiv:2002.03734.

\bibitem[{Feng, wen Gao, and Luo(2021)}]{8}
Feng, X.; wen Gao, X.; and Luo, L. 2021.
\newblock X-SDD: A New Benchmark for Hot Rolled Steel Strip Surface Defects Detection.
\newblock \emph{Symmetry}, 13: 706.

\bibitem[{Gong et~al.(2019)Gong, Liu, Le, Saha, Mansour, Venkatesh, and van~den Hengel}]{30}
Gong, D.; Liu, L.; Le, V.; Saha, B.; Mansour, M.~R.; Venkatesh, S.; and van~den Hengel, A. 2019.
\newblock Memorizing Normality to Detect Anomaly: Memory-augmented Deep Autoencoder for Unsupervised Anomaly Detection.
\newblock arXiv:1904.02639.

\bibitem[{Guo et~al.(2020)Guo, Li, Jiang, and Shen}]{10}
Guo, L.; Li, R.; Jiang, B.; and Shen, X. 2020.
\newblock Automatic crack distress classification from concrete surface images using a novel deep-width network architecture.
\newblock \emph{Neurocomputing}, 397: 383--392.

\bibitem[{Hou et~al.(2021)Hou, Zhang, Zhong, Xie, Pu, and Zhou}]{31}
Hou, J.; Zhang, Y.; Zhong, Q.; Xie, D.; Pu, S.; and Zhou, H. 2021.
\newblock Divide-and-Assemble: Learning Block-wise Memory for Unsupervised Anomaly Detection.
\newblock arXiv:2107.13118.

\bibitem[{Kingma and Welling(2022)}]{23}
Kingma, D.~P.; and Welling, M. 2022.
\newblock Auto-Encoding Variational Bayes.
\newblock arXiv:1312.6114.

\bibitem[{Lei et~al.(2023)Lei, Hu, Wang, and Liu}]{47}
Lei, J.; Hu, X.; Wang, Y.; and Liu, D. 2023.
\newblock PyramidFlow: High-Resolution Defect Contrastive Localization using Pyramid Normalizing Flow.
\newblock arXiv:2303.02595.

\bibitem[{Liu, Wu, and Lv(2023)}]{15}
Liu, Y.; Wu, G.; and Lv, Z. 2023.
\newblock SDGAN: A novel spatial deformable generative adversarial network for low-dose CT image reconstruction.
\newblock \emph{Displays}, 78: 102405.

\bibitem[{Liu et~al.(2023)Liu, Zhou, Xu, and Wang}]{39}
Liu, Z.; Zhou, Y.; Xu, Y.; and Wang, Z. 2023.
\newblock SimpleNet: A Simple Network for Image Anomaly Detection and Localization.
\newblock arXiv:2303.15140.

\bibitem[{Ninja(2024)}]{1}
Ninja, D. 2024.
\newblock Visualization Tools for Industrial Optical Inspection Dataset.
\newblock \url{ https://datasetninja.com/industrial-optical-inspection }.
\newblock Visited on 2024-08-14.

\bibitem[{Niu et~al.(2021)Niu, Song, Huang, Wang, Yan, and Meng}]{7}
Niu, M.; Song, K.; Huang, L.; Wang, Q.; Yan, Y.; and Meng, Q. 2021.
\newblock Unsupervised Saliency Detection of Rail Surface Defects Using Stereoscopic Images.
\newblock \emph{IEEE Transactions on Industrial Informatics}, 17(3): 2271--2281.

\bibitem[{Park, Noh, and Ham(2020)}]{32}
Park, H.; Noh, J.; and Ham, B. 2020.
\newblock Learning Memory-guided Normality for Anomaly Detection.
\newblock arXiv:2003.13228.

\bibitem[{Perera, Nallapati, and Xiang(2019)}]{25}
Perera, P.; Nallapati, R.; and Xiang, B. 2019.
\newblock OCGAN: One-Class Novelty Detection Using GANs With Constrained Latent Representations.
\newblock In \emph{2019 IEEE/CVF Conference on Computer Vision and Pattern Recognition (CVPR)}, 2893--2901.

\bibitem[{Pourreza et~al.(2020)Pourreza, Mohammadi, Khaki, Bouindour, Snoussi, and Sabokrou}]{35}
Pourreza, M.; Mohammadi, B.; Khaki, M.; Bouindour, S.; Snoussi, H.; and Sabokrou, M. 2020.
\newblock G2D: Generate to Detect Anomaly.
\newblock arXiv:2006.11629.

\bibitem[{Reiss and Hoshen(2022)}]{46}
Reiss, T.; and Hoshen, Y. 2022.
\newblock Mean-Shifted Contrastive Loss for Anomaly Detection.
\newblock arXiv:2106.03844.

\bibitem[{Rippel, Müller, and Merhof(2020)}]{13}
Rippel, O.; Müller, M.; and Merhof, D. 2020.
\newblock GAN-based Defect Synthesis for Anomaly Detection in Fabrics.
\newblock In \emph{2020 25th IEEE International Conference on Emerging Technologies and Factory Automation (ETFA)}, volume~1, 534--540.

\bibitem[{Sabokrou et~al.(2018)Sabokrou, Khalooei, Fathy, and Adeli}]{20}
Sabokrou, M.; Khalooei, M.; Fathy, M.; and Adeli, E. 2018.
\newblock Adversarially Learned One-Class Classifier for Novelty Detection.
\newblock \emph{CoRR}, abs/1802.09088.

\bibitem[{Salehi et~al.(2020)Salehi, Sadjadi, Baselizadeh, Rohban, and Rabiee}]{38}
Salehi, M.; Sadjadi, N.; Baselizadeh, S.; Rohban, M.~H.; and Rabiee, H.~R. 2020.
\newblock Multiresolution Knowledge Distillation for Anomaly Detection.
\newblock arXiv:2011.11108.

\bibitem[{Shi, Yang, and Qi(2021)}]{28}
Shi, Y.; Yang, J.; and Qi, Z. 2021.
\newblock Unsupervised anomaly segmentation via deep feature reconstruction.
\newblock \emph{Neurocomputing}, 424: 9--22.

\bibitem[{Silvestre-Blanes et~al.(2019)Silvestre-Blanes, Albero-Albero, Miralles, Pérez-Llorens, and Moreno}]{2}
Silvestre-Blanes, J.; Albero-Albero, T.; Miralles, I.; Pérez-Llorens, R.; and Moreno, J. 2019.
\newblock A Public Fabric Database for Defect Detection Methods and Results.
\newblock \emph{Autex Research Journal}, 19.

\bibitem[{Song, Song, and Yan(2020)}]{5}
Song, G.; Song, K.; and Yan, Y. 2020.
\newblock Saliency detection for strip steel surface defects using multiple constraints and improved texture features.
\newblock \emph{Optics and Lasers in Engineering}, 128: 106000.

\bibitem[{Texture-ad(2024)}]{50}
Texture-ad. 2024.
\newblock Texture-AD-Benchmark.
\newblock \url{https://huggingface.co/datasets/texture-ad/Texture-AD-Benchmark}.
\newblock Accessed: 2024-08-15.

\bibitem[{Wang et~al.(2021)Wang, Han, Ding, and Huang}]{37}
Wang, G.; Han, S.; Ding, E.; and Huang, D. 2021.
\newblock Student-Teacher Feature Pyramid Matching for Anomaly Detection.
\newblock arXiv:2103.04257.

\bibitem[{Wu, Jang, and Chen(2015)}]{3}
Wu, M.-J.; Jang, J.-S.~R.; and Chen, J.-L. 2015.
\newblock Wafer Map Failure Pattern Recognition and Similarity Ranking for Large-Scale Data Sets.
\newblock \emph{IEEE Transactions on Semiconductor Manufacturing}, 28(1): 1--12.

\bibitem[{Xia et~al.(2020)Xia, Zhang, Liu, Shen, and Yuille}]{29}
Xia, Y.; Zhang, Y.; Liu, F.; Shen, W.; and Yuille, A. 2020.
\newblock Synthesize then Compare: Detecting Failures and Anomalies for Semantic Segmentation.
\newblock arXiv:2003.08440.

\bibitem[{Xu et~al.(2019)Xu, Su, Wang, Cai, Cui, and Chen}]{11}
Xu, H.; Su, X.; Wang, Y.; Cai, H.; Cui, K.; and Chen, X. 2019.
\newblock Automatic Bridge Crack Detection Using a Convolutional Neural Network.
\newblock \emph{Applied Sciences}, 9: 2867.

\bibitem[{Yan et~al.(2021)Yan, Zhang, Xu, Hu, and Heng}]{34}
Yan, X.; Zhang, H.; Xu, X.; Hu, X.; and Heng, P.-A. 2021.
\newblock Learning Semantic Context from Normal Samples for Unsupervised Anomaly Detection.
\newblock In \emph{AAAI Conference on Artificial Intelligence}.

\bibitem[{You et~al.(2022)You, Yang, Luo, Cui, Zheng, and Le}]{21}
You, Z.; Yang, K.; Luo, W.; Cui, L.; Zheng, Y.; and Le, X. 2022.
\newblock ADTR: Anomaly Detection Transformer with Feature Reconstruction.
\newblock arXiv:2209.01816.

\bibitem[{Zaheer et~al.(2020)Zaheer, ha~Lee, Astrid, and Lee}]{26}
Zaheer, M.~Z.; ha~Lee, J.; Astrid, M.; and Lee, S.-I. 2020.
\newblock Old is Gold: Redefining the Adversarially Learned One-Class Classifier Training Paradigm.
\newblock arXiv:2004.07657.

\bibitem[{Zavrtanik, Kristan, and Skocaj(2021)}]{16}
Zavrtanik, V.; Kristan, M.; and Skocaj, D. 2021.
\newblock DR{\AE}M - {A} discriminatively trained reconstruction embedding for surface anomaly detection.
\newblock \emph{CoRR}, abs/2108.07610.

\bibitem[{Zhang, Wang, and Kuo(2021)}]{19}
Zhang, K.; Wang, B.; and Kuo, C.~J. 2021.
\newblock PEDENet: Image Anomaly Localization via Patch Embedding and Density Estimation.
\newblock \emph{CoRR}, abs/2110.15525.

\bibitem[{Zhang et~al.(2021)Zhang, Yu, Yang, Zhou, and Zhao}]{9}
Zhang, Z.; Yu, S.; Yang, S.; Zhou, Y.; and Zhao, B. 2021.
\newblock Rail-5k: a Real-World Dataset for Rail Surface Defects Detection.
\newblock \emph{CoRR}, abs/2106.14366.

\bibitem[{Zhao et~al.(2022)Zhao, Shi, Yin, Dong, Zhang, Kang, Yu, Chen, Li, Liu, and Zhang}]{6}
Zhao, X.; Shi, J.; Yin, Q.; Dong, Z.; Zhang, Y.; Kang, L.; Yu, Q.; Chen, C.; Li, J.; Liu, X.; and Zhang, K. 2022.
\newblock Controllable synthesis of high-quality two-dimensional tellurium by a facile chemical vapor transport strategy.
\newblock \emph{iScience}, 25(1): 103594.

\bibitem[{Zhou et~al.(2020)Zhou, Xiao, Yang, Cheng, Liu, Luo, Gu, Liu, and Gao}]{27}
Zhou, K.; Xiao, Y.; Yang, J.; Cheng, J.; Liu, W.; Luo, W.; Gu, Z.; Liu, J.; and Gao, S. 2020.
\newblock Encoding Structure-Texture Relation with P-Net for Anomaly Detection in Retinal Images.
\newblock In Vedaldi, A.; Bischof, H.; Brox, T.; and Frahm, J.-M., eds., \emph{Computer Vision -- ECCV 2020}, 360--377. Cham: Springer International Publishing.
\newblock ISBN 978-3-030-58565-5.

\bibitem[{Zhou et~al.(2023)Zhou, Xu, Song, Shen, and Shen}]{49}
Zhou, Y.; Xu, X.; Song, J.; Shen, F.; and Shen, H.~T. 2023.
\newblock MSFlow: Multi-Scale Flow-based Framework for Unsupervised Anomaly Detection.
\newblock arXiv:2308.15300.

\end{thebibliography}
\end{document}